# Perturbative Neural Networks

Felix Juefei-Xu
Carnegie Mellon University
felixu@cmu.edu

Vishnu Naresh Boddeti
Michigan State University
vishnu@msu.edu

Marios Savvides
Carnegie Mellon University
msavvid@ri.cmu.edu

## Abstract

*Convolutional neural networks are witnessing wide adoption in computer vision systems with numerous applications across a range of visual recognition tasks. Much of this progress is fueled through advances in convolutional neural network architectures and learning algorithms even as the basic premise of a convolutional layer has remained unchanged. In this paper, we seek to revisit the convolutional layer that has been the workhorse of state-of-the-art visual recognition models. We introduce a very simple, yet effective, module called a perturbation layer as an alternative to a convolutional layer. The perturbation layer does away with convolution in the traditional sense and instead computes its response as a weighted linear combination of non-linearly activated additive noise perturbed inputs. We demonstrate both analytically and empirically that this perturbation layer can be an effective replacement for a standard convolutional layer. Empirically, deep neural networks with perturbation layers, called **Perturbative Neural Networks (PNNs)**, in lieu of convolutional layers perform comparably with standard CNNs on a range of visual datasets (MNIST, CIFAR-10, PASCAL VOC, and ImageNet) with fewer parameters.*

## 1. Introduction

Deep convolutional neural networks (CNNs) have been overwhelmingly successful across a variety of visual perception tasks. The past several years have witnessed the evolution of many successful CNN architectures such as AlexNet [14], VGG [27], GoogLeNet [30], ResNet [8, 9], MobileNet [10], and DenseNet [11], *etc*. Much of this effort has been focused on the topology and connectivity between convolutional and other modules while the convolutional layer itself has continued to remain the backbone of these networks. Convolutional layers are characterized by two main properties [17, 5], namely, local connectivity and weight sharing, both of which afford these layers with significant computational and statistical efficiency over densely connected layers. Ever since the introduction of AlexNet [14], there has been steady refinements to a standard convolutional layer. While AlexNet utilized convolutional filters with large receptive fields ($11 \times 11$, $5 \times 5$, *etc*.), the VGG network [27] demonstrated the utility of using convolutional weights with very small receptive fields ($3 \times 3$) that are both statistically and computationally efficient for learning deep convolutional neural networks. As convolutional layers are often the main computational bottleneck of CNNs, there has been steady developments in improving the computational efficiency of convolutional layers. MobileNets [10] introduced efficient reparameterization of standard $3 \times 3$ convolutional weights, in terms of depth-wise convolutions and $1 \times 1$ convolutions. Convolutional networks with binary weights [2, 3, 23] have been proposed to significantly improve the computational efficiency of CNNs. Recent work has also demonstrated that sparse convolutional weights [20, 21, 18] perform comparably to dense convolutional weights while also being computationally efficient. However, across this entire body of work the basic premise of a convolutional layer itself has largely remained unchanged.

This paper seeks to rethink the basic premise of the necessity of convolutional layers for the task of image classification. The success of a wide range of approaches that utilize convolutional layers that have, a) very small receptive fields ($3 \times 3$), b) sparse convolutional weights, and c) convolutional weights with binary weights, motivates our hypothesis that one can perhaps completely do away with convolutional layers for learning high performance image classification models. We propose a novel module, dubbed the perturbation layer[1], that conforms to our hypothesis and is devoid of standard convolutional operations. Given an input, the perturbation layer first perturbs the input additively through random, but fixed, noise followed by a weighted combination of non-linear activations of the input perturbations. The weighted linear combinations of activated perturbations are conceptually similar to $1 \times 1$ convolutions, but are not strictly convolutional since their receptive field is just one pixel, as opposed to the receptive fields of standard convolutional weights. This layer is thus an extreme version of sparse convolutional weights with sparsity of one non-zero element and fixed non-zero support at the center of the filter. Avoiding convolutions with receptive fields larger than one offers immediate statistical savings in the form of fewer learnable network parameters, computational savings from more efficient operations (weighted sum vs. convolution) and more importantly allows us to rethink the premise and utility of convolutional layers in the context of image classification models. Our theoretical analysis shows that the perturbation layer can approximate the response of a standard convolutional layer. In addition, we empirically demonstrate that deep neural networks with the perturbation layers as replacements for standard convolutional layers perform as well as an equivalent network with convolutional layers across a variety of datasets of varying difficulty and scale, MNIST, CIFAR-10, PASCAL VOC, and ImageNet.

---

[1] Implementation and future updates will be available at http://xujuefei.com/pnn.

## 2. Related Work

There is a huge body of work on the design and applications of CNNs for image classification, the full treatment of which is beyond the scope of this paper. We will however note a few major advances that were motivated by improving the performance of these networks, such as, AlexNet [14], VGG [27], GoogLeNet [30], Residual Networks [8], *etc*.

The idea of using convolutional weights with small receptive fields is not new. While the VGG [27] network was the first model to demonstrate the efficacy of small convolutional weights in deep CNNs, other researchers have explored the use of small convolutional weights, including $1 \times 1$ convolutional weights. For instance the GoogLeNet [30] architecture comprises of weights with different receptive fields including $1 \times 1$ weights. The Network in Network architecture [19] also utilizes $1 \times 1$ convolutions. However, all of these approaches have used $1 \times 1$ convolutions in conjunction with convolutional filters with larger receptive fields. In contrast the perturbative layer that we introduce in this paper is devoid of any convolutional layers with receptive fields larger than one pixel and combines information from multiple noise perturbed versions of the input.

Efficient characterization of convolutional layers have also been proposed from the perspective of computational efficiency. Networks with binary weights [3, 2, 23], networks with sparse convolutional weights [20, 21, 18], networks with efficient factorization of the convolutional weights [10, 15] and networks with a hybrid of learnable and fixed weights [12]. While the proposed perturbation layer does offer computational benefits in terms of fewer parameters and more efficient inference our aim in this paper is to motivate the need to rethink the premise and utility of a standard convolutional layer for image classification tasks. The proposed perturbation layer serves as evidence that perhaps convolutional layers are not very critical to learning image classification models that can perform as well as, if not better than, standard convolutional networks.

## 3. Proposed Method

In this section, we first detail the motivation and formulation of the proposed perturbative neural networks (PNN), and then discuss its relation to standard CNNs from both a macro as well as a micro viewpoint. Finally, we discuss some properties associated with PNN.

### 3.1. Revisiting LBCNN and an Observation

Recently, local binary convolutional neural networks (LBCNN) (as in [12]) have been motivated from the local binary patterns (LBP) descriptor. The basic LBCNN module is shown in the middle row of Figure 1, where the input image (or tensor at subsequent layers) is first convolved with a set of fixed, randomly generated sparse binary filters, and the resulting response map is propagated through a nonlinear activation, such as ReLU, and then, the ReLU activated response map is linearly combined to generate an output feature map that feeds into the next layer. These linear combination weights are the only learnable parameters.

Specifically, let us assume that the input image (or tensor) $\mathbf{x}_l$ is filtered by $m$ pre-defined fixed binary filters $\mathbf{b}_i, i \in [m]$, to generate $m$ difference maps that are then activated through ReLU, resulting in $m$ response maps. The linear combination weights for the $m$ response maps are $\mathcal{W}_{l,i}, i \in [m]$ for obtaining one final feature map. The combined set of feature maps serve as the input $\mathbf{x}_{l+1}$ for the next layer.

Then the transfer function between input and output of layer $l$ can be expressed as:

$$\mathbf{x}_{l+1}^t = \sum_{i=1}^{m} \sigma_{\text{relu}} \left( \sum_s \mathbf{b}_{l,i}^s * \mathbf{x}_l^s \right) \cdot \mathcal{W}_{l,i}^t \quad (1)$$

where $t$ is the output channel, $s$ is the input channel, and $*$ is the channel-wise convolution operation. Again, linear weights $\mathcal{W}$ are the only learnable parameters of an LBCNN layer. In this way, LBCNN can yield much lower model complexity with significant savings in the number of learnable parameters.

LBCNN's idea of formulating a deep learning model with a hybrid of fixed convolutional weights and learnable linear combination weights in each layer is intriguing. In Figure 2, we can see that a $3 \times 3$ patch from the kitten image is extracted, and the pixels are labeled as $x_1, \ldots, x_9$. This patch, in LBCNN, will first be convolved with a binary filter, which is shown on the far right as an example. Since the filter itself is binary with $+1$ and $-1$, the resulting scalar on the response map will simply be the additions and subtractions among the 9 neighboring pixels. In this case, it reduces to $y = x_1 + x_3 + x_5 + x_7 + x_9 - x_2 - x_4 - x_6 - x_8$. The same process repeats for the next $3 \times 3$ patch until the entire response map is generated.

Mathematically, this convolution operation is transforming the patch center pixel $x_c = x_5$ to one particular point $y$

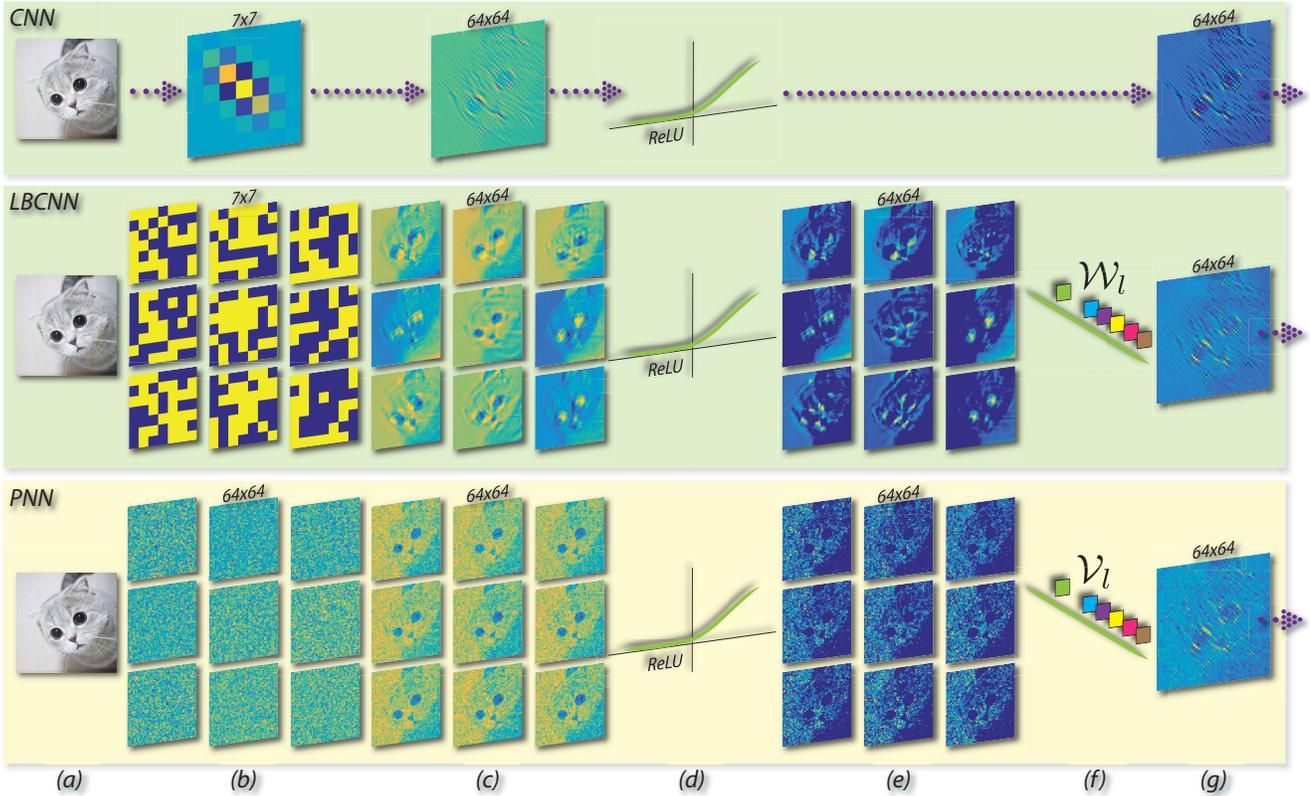

**Figure 1:** Basic modules in CNN, LBCNN [12], and PNN. $\mathcal{W}_l$ and $\mathcal{V}_l$ are the learnable weights for local binary convolution layer and the proposed perturbation layer respectively. Inspired by the formulation of LBCNN, the proposed PNN method also uses a set of linear weights to combine various perturbation maps. For CNN: (a) input, (b) learnable convolutional filter, (c) response map, (d) ReLU, (g) feature map. For LBCNN: (a) input, (b) fixed non-learnable binary filters, (c) difference maps by convolving with binary filters, (d) ReLU, (e) activated difference maps, (f) learnable linear weights for combining the activated difference maps, (g) feature map. For PNN: (a) input, (b) fixed non-learnable perturbation masks, (c) response maps by addition with perturbation masks, (d) ReLU, (e) activated response maps, (f) learnable linear weights for combining the activated response maps, (g) feature map.

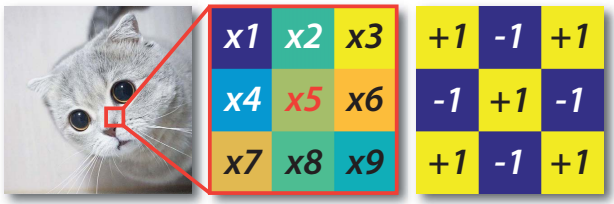

**Figure 2:** Convolving with a binary filter is equivalent to addition and subtraction among neighbors within the patch. Similarly, convolving with a real-valued filter is equivalent to the linear combination of the neighbors using filter weights.

on the response map: $y = f(x_c)$, and in this case, the function $f$ is defined as above by including 8 neighboring pixels of $x_c$ as well as the binary filter itself. which determines the pixels that are added or subtracted. The same notion can be easily extended towards standard convolution operations with real-valued convolutional filters, either learnable or non-learnable, and then the additions and subtractions among the neighboring pixels become a linear weighted combinations.

This seemingly simple inner product operation $f$, whether it has been efficiently implemented in the frequency domain or spatial domain, is a major computational bottleneck of deep CNN models. The key takeaway concept from LBCNN is that the convolutional weights could be made non-learnable and the learning can be carried out solely through the linear combination weights, suggesting that there may be more potential simplifications to the spatial convolution layer, since the primary network optimization happens through the linear combination weights. This begs the question, can we build upon LBCNN and arrive at a much simpler function $\hat{f}$?

### 3.2. Perturbative Neural Networks Module

The aforementioned observation motivates the formulation of the PNN. The pictorial illustration of the proposed PNN module is shown in the bottom row of Figure 1. When the input image comes in, it will be perturbed with a set of pre-defined random additive noise masks, each with the same size as the input image, resulting in a set of noise-perturbed maps. These maps will go through a ReLU non-linearity and are then linearly combined to form one final feature map. Again, the random additive noise masks are pre-defined and non-learnable, and the only learnable parameters are the linear combination weights. Mathematically, PNN transforms the input and output of layer $l$ in the following way:

$$\mathbf{x}_{l+1}^t = \sum_{i=1}^{m} \sigma_{\text{relu}}\left(\mathcal{N}_l^i + \mathbf{x}_l^i\right) \cdot \mathcal{V}_{l,i}^t \quad (2)$$

where $t$ is the output channel index, $i$ is the input channel index, and $\mathcal{N}_l^i$ is the $i$-th random additive perturbation mask in layer $l$. Similar to LBCNN, the linear weights $\mathcal{V}$ are the only learnable parameters of a perturbation layer. Not surprisingly, PNN is able to save lots of learnable parameters as will be discussed in the following sections.

From Eq. 2 we observe that the computationally expensive convolution operation is replaced by an element-wise noise addition which is significantly more efficient. Recall in the previous section we ask the question whether it is possible to arrive at a much simpler function $\hat{f}$ that transforms the patch center $x_c$ to one point $y$ on the feature map. Now we can have $y = f(x_c) = x_c + n_c$, where $n_c$ is the added noise corresponding to $x_c$ location. An attractive attribute of the PNN formulation is that repetitive operation such as the convolution (moving from one patch to the other) is no longer needed. A single pass for adding the noise perturbation mask to the entire input channel completes the task.

### 3.3. Relating PNN and CNN: A Macro View

Let $\mathbf{x} \in \mathbb{R}^d$ be a vectorized input image of dimension $d$ and let $\mathbf{y} \in \mathbb{R}^d$ be a vectorized feature map after convolving $\mathbf{x}$ with a 2D convolutional filter $\mathbf{w} \in \mathbb{R}^{k \times k}$. We use the notation $\text{vec}(\cdot)$ to represent the vectorization of a matrix and $\text{mat}(\cdot)$ to represent the opposite, which is reshaping the vector to its original matrix form. The following discussion will be done by using 2D matrices but the same technique applies for high-dimensional tensors as practiced in CNN layers as well. Therefore, the standard CNN convolution operation is as follows, assuming no bias in the convolution:

$$\text{CNN}: \quad \mathbf{y} = \text{vec}(\text{mat}(\mathbf{x}) * \mathbf{w}) = \sum_{i}^{k^2} \mathbf{x}_{i,\text{shift}} \cdot w_i \quad (3)$$

where $\mathbf{x}_{i,\text{shift}}$ is the $i$-th spatially shifted version of the input in vectorized form, and $w_i$ is the $i$-th element in the convolution filter $\mathbf{w}$.

For PNN, the same input $\mathbf{x}$ will be perturbed with $N$ random noise masks $\mathbf{n}_i$, and then linearly combined using weight vector $\mathbf{v}$ whose elements are $v_i$'s to form the output response vector $\hat{\mathbf{y}}$. Therefore, for PNN, the operation follows:

$$\text{PNN}: \quad \hat{\mathbf{y}} = \sum_{i=1}^{N} (\mathbf{x} + \mathbf{n}_i) \cdot v_i \quad (4)$$

If we arrange $\mathbf{x} + \mathbf{n}_i$ as the columns vectors of a matrix $\hat{\mathbf{X}} \in \mathbb{R}^{d \times N}$, we can rewrite the PNN operation as:

$$\hat{\mathbf{y}} = \hat{\mathbf{X}}\mathbf{v} = (\mathbf{X} + \mathbf{N})\mathbf{v} = (\mathbf{x}\mathbf{1}^\top + \mathbf{N})\mathbf{v} \quad (5)$$

where $\mathbf{X} \in \mathbb{R}^{d \times N}$ has vector $\mathbf{x}$ repeated in its columns and $\mathbf{N} \in \mathbb{R}^{d \times N}$ is a perturbation matrix with noise vector $\mathbf{n}_i$ in its columns.

Given the CNN output vector $\mathbf{y}$, we can always find the vector $\mathbf{v}$ for PNN such that the PNN output $\hat{\mathbf{y}}$ is equal to or approximately equal to $\mathbf{y}$. If $N = d$, $\hat{\mathbf{X}}$ is a full rank square matrix so an exact solution for $\mathbf{v}$ can be found as:

$$\mathbf{v}^* = \hat{\mathbf{X}}^{-1}\mathbf{y} = (\mathbf{x}\mathbf{1}^\top + \mathbf{N})^{-1}\mathbf{y} \quad (6)$$

$$= \left[\mathbf{N}^{-1} - \frac{\mathbf{N}^{-1}\mathbf{x}\mathbf{1}^\top \mathbf{N}^{-1}}{1 + \mathbf{1}^\top \mathbf{N}^{-1}\mathbf{x}}\right]\mathbf{y} \quad (7)$$

where the last step is due to Sherman–Morrison formula [25, 26, 22]. If $N < d$, $\hat{\mathbf{X}}$ is a tall matrix, so a least square solution can be found for $\mathbf{v}$ as:

$$\mathbf{v}^* = (\hat{\mathbf{X}}^\top \hat{\mathbf{X}})^{-1}\hat{\mathbf{X}}^\top \mathbf{y} \quad (8)$$

Next, we will derive a relationship between the convolutional weights in CNN and the perturbation weights in PNN assuming $\hat{\mathbf{y}} = \mathbf{y}$. Recall that convolution is a linear operation that transforms input $\mathbf{x}$ to output $\mathbf{y}$ and can be viewed as multiplication of a matrix. So we can rewrite the convolution operation simply as:

$$\mathbf{y} = \mathbf{A}\mathbf{x} \quad (9)$$

where $\mathbf{A}$ is a doubly block circulant matrix which corresponds to convolutional weights $\mathbf{w}$ with proper manipulation. Using the derived optimal linear weights vector $\mathbf{v}^*$, the PNN reconstruction simplifies to:

$$\hat{\mathbf{y}}_r = (\mathbf{x}\mathbf{1}^\top + \mathbf{N})\mathbf{v}^* \quad (10)$$

$$= (\mathbf{x}\mathbf{1}^\top + \mathbf{N})\left[\mathbf{N}^{-1} - \frac{\mathbf{N}^{-1}\mathbf{x}\mathbf{1}^\top \mathbf{N}^{-1}}{1 + \mathbf{1}^\top \mathbf{N}^{-1}\mathbf{x}}\right]\mathbf{y} \quad (11)$$

$$= \mathbf{A}\mathbf{x} = \mathbf{y} \quad (12)$$

Therefore, we can establish the following relationship:

$$\Rightarrow (\mathbf{x}\mathbf{1}^\top + \mathbf{N})\left[\mathbf{N}^{-1}\mathbf{y} - \frac{\mathbf{N}^{-1}\mathbf{x}\mathbf{1}^\top \mathbf{N}^{-1}\mathbf{y}}{1 + \mathbf{1}^\top \mathbf{N}^{-1}\mathbf{x}}\right] = \mathbf{A}\mathbf{x} \quad (13)$$

$$\Rightarrow \mathbf{x}\underbrace{\mathbf{1}^\top \mathbf{N}^{-1}}_{\mathbf{n}^\top}(\mathbf{A}\mathbf{x})\underbrace{\mathbf{1}^\top \mathbf{N}^{-1}}_{\mathbf{n}^\top}\mathbf{x} = \mathbf{x}\underbrace{\mathbf{1}^\top \mathbf{N}^{-1}}_{\mathbf{n}^\top}\mathbf{x}\underbrace{\mathbf{1}^\top \mathbf{N}^{-1}}_{\mathbf{n}^\top}(\mathbf{A}\mathbf{x})$$

$$\Rightarrow \mathbf{x}\mathbf{n}^\top(\mathbf{A}\mathbf{x}\mathbf{n}^\top)\mathbf{x} = \mathbf{x}\mathbf{n}^\top(\mathbf{x}\mathbf{n}^\top \mathbf{A})\mathbf{x} \quad (14)$$

By observation, the following must hold:

$$\mathbf{A}\mathbf{x}\mathbf{n}^\top = \mathbf{x}\mathbf{n}^\top \mathbf{A} \quad (15)$$

$$\Rightarrow \mathbf{A}\mathbf{X}\mathbf{N}^{-1} = \mathbf{X}\mathbf{N}^{-1}\mathbf{A} \quad (16)$$

$$\Rightarrow (\mathbf{X}^+ \mathbf{A}\mathbf{X})\mathbf{N}^{-1} = \mathbf{N}^{-1}\mathbf{A} \quad (17)$$

where $\mathbf{X}^+$ is the Moore–Penrose inverse of $\mathbf{X}$. Rearranging the terms, we can arrive at the Sylvester equation [29] commonly used in control theory:

$$\underbrace{(\mathbf{X}^+\mathbf{A}\mathbf{X})}_{\mathcal{S}_a}\mathbf{N}^{-1} + \mathbf{N}^{-1}\underbrace{(-\mathbf{A})}_{\mathcal{S}_b} = \underbrace{\mathbf{0}}_{\mathcal{S}_c} \quad (18)$$

Reformulating in terms of Kronecker tensor product we have:

$$\left[\mathbf{I} \otimes \mathcal{S}_a + \mathcal{S}_b^\top \otimes \mathbf{I}\right]\mathbf{N}^{-1}(:) = \mathcal{S}_c(:) \quad (19)$$

in that $\mathbf{N}^{-1}$ will have a unique solution when the eigenvalues of $\mathcal{S}_a$ and $-\mathcal{S}_b$ are distinct, meaning the spectra of $(\mathbf{X}^+\mathbf{A}\mathbf{X})$ and $\mathbf{A}$ are disjoint. In this way, given the known input $\mathbf{x}$ and convolution transformation matrix $\mathbf{A}$, we can always solve for the matching noise perturbation matrix $\mathbf{N}$ using linear algebra toolbox such as the Matlab Sylvester equation routine.

### 3.4. Relating PNN and CNN: A Micro View

Now let us consider a single neighborhood (patch) in the input tensor where the convolution is taking place, and obtain a relation between PNN and CNN with some mild assumptions. Let us assume that each pixel $x_i$ within this patch is a random variable and we call the central pixel $x_c$ for simplicity which has a total of $\text{card}(\mathbb{N}_c)$ neighbors where $\mathbb{N}_c$ is a set containing the indices of the neighboring pixels of $x_c$. Let us further make assumptions on the first and second order statistics of $x_i$. In this case, we assume that $E(x_i) = 0$ and $E(x_i^2) = \sigma^2$. Let $\epsilon_i = x_i - x_c$, $i \in \mathbb{N}_c$ be the difference between neighbor $x_i$ and the central pixel $x_c$. Next we want to examine the following three quantities, namely $E(\epsilon_i)$, $E(\epsilon_i^2)$, and $E(\epsilon_i \epsilon_j)$, which will be used for the subsequent derivation.

First, it is quite easy to see that: $E(\epsilon_i) = E(x_i - x_c) = 0$. Next, for the second order statistics $E(\epsilon_i^2)$, we have:

$$\begin{aligned}E(\epsilon_i^2) &= E[(x_i - x_c)^2] = E(x_i^2 + x_c^2 - 2x_i x_c) \\ &= E(x_i^2) + E(x_c^2) - 2E(x_i x_c) \\ &= 2\sigma^2 - 2\rho\sigma^2 = 2\sigma^2\delta\end{aligned} \quad (20)$$

where $\delta = 1 - \rho$. In this case, we assume that $\rho \approx 1$ because neighboring pixels usually have high correlations. Therefore, $\delta$ is usually very small meaning that $E(\epsilon_i^2)$ is very small as well. Lastly, for $E(\epsilon_i \epsilon_j)$, $i \neq j$ we have:

$$\begin{aligned}E(\epsilon_i \epsilon_j) &= E[(x_i - x_c)(x_j - x_c)] \\ &= E(x_c^2) - E(x_i x_c) - E(x_j x_c) + E(x_i x_j) \\ &= \sigma^2 - \rho\sigma^2 - \rho\sigma^2 + \hat{\rho}\sigma^2 \text{ (assuming } \hat{\rho} \approx \rho) \\ &= \sigma^2 - \rho\sigma^2 = \sigma^2\delta = (1/2)E(\epsilon_i^2)\end{aligned} \quad (21)$$

For CNN, the convolution operation maps the central pixel $x_c$ to one point $y$ on the output feature map with convolutional weights $w_i$'s as follows. Let $N = \text{card}(\mathbb{N}_c) + 1$:

$$y = \sum_{i=1}^{N} x_i w_i = x_c + \sum_{i \in \mathbb{N}_c} x_i w_i \quad (22)$$

$$\Rightarrow x_c w_c + \sum_{i \in \mathbb{N}_c}(x_c + \epsilon_i) w_i = y \quad (23)$$

$$\Rightarrow x_c\left(w_c + \sum_{i \in \mathbb{N}_c} w_i\right) + \sum_{i \in \mathbb{N}_c} \epsilon_i w_i = y \quad (24)$$

$$\Rightarrow x_c + \sum_{i \in \mathbb{N}_c} \epsilon_i \left(\frac{w_i}{\sum_i^N w_i}\right) = \frac{y}{\sum_i^N w_i} \quad (25)$$

$$\Rightarrow x_c + \underbrace{\sum_{i \in \mathbb{N}_c} \epsilon_i w_i'}_{n_c} = y' \quad (26)$$

Establishing that $n_c = \sum_{i \in \mathbb{N}_c} \epsilon_i w_i'$ behaves like additive perturbation noise, will allows us to relate the CNN forumulation to the PNN formulation.

Next, we will examine $E(n_c)$ and $E(n_c^2)$. First, it can be easily shown that $E(n_c) = E\left(\sum_{i \in \mathbb{N}_c} \epsilon_i w_i'\right) = 0$ since $E(\epsilon_i) = 0$. Next, for the second order statistics, we have:

$$\begin{aligned}E(n_c^2) &= E\left(\sum_{i \in \mathbb{N}_c} \epsilon_i w_i'\right)^2 = E\left(\sum_{i=1}^N \epsilon_i w_i'\right)^2 \text{ since } \epsilon_c = 0 \\ &= E(\epsilon_1^2 w_1'^2 + \ldots \epsilon_i^2 w_i'^2 + \ldots + \underbrace{\epsilon_1\epsilon_2 w_1' w_2' + \ldots}_{\text{cross-terms}}) \\ &= E(\epsilon_i^2)\sum_{i=1}^N w_i'^2 + E(\epsilon_i \epsilon_j)\sum_i \sum_{j \neq i} w_i' w_j' \\ &= (2\sigma^2\delta)\|\mathbf{w}'\|_2^2 + (\sigma^2\delta)\sum_i \sum_{j \neq i} w_i' w_j' \\ &= 2\sigma^2\delta\left[\|\mathbf{w}'\|_2^2 + (1/2)\sum_i \sum_{j \neq i} w_i' w_j'\right] \\ &= 2\sigma^2\delta' \text{ (small)}\end{aligned} \quad (27)$$

where $\delta' = \delta[\|\mathbf{w}'\|_2^2 + (1/2)\sum_i \sum_{j \neq i} w_i' w_j']$. Therefore, this analysis of the CNN operation establishes a relation between the CNN and the PNN formulation. However, one may notice that in Eq. 26, the RHS is $y'$ instead of $y$. By allowing multiple perturbation maps to combine using the linear combination weights as shown before leads to $y$ on the RHS.

### 3.5. Properties of PNN

Recall that convolution leverages two important ideas that can help improve a machine learning system: sparse interactions and parameter sharing [5]. Not surprisingly, the proposed PNN also share many of these nice properties.

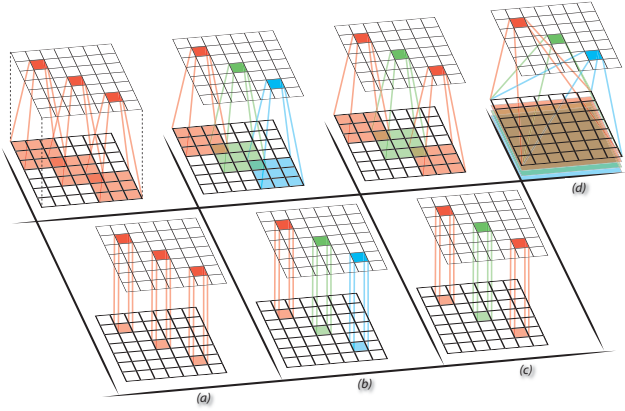

**Figure 3:** Variations in connectivity patterns among commonly practiced types of convolutional operations such as (a) the regular convolution, (b) locally connected convolution, (c) tiled convolution, and finally (d) fully connected layer. For (a-c), top row is a $3 \times 3$ convolution and the bottom row is a $1 \times 1$ convolution.

**Sparse interactions**: Firstly, PNN adds perturbation to the input with perturbation mask of the same size as the input. Therefore, it is easy to see that it only needs a single element in the input to contribute to one element in the output perturbation map. Hence, sparse interaction. Secondly, PNN utilizes a set of learnable linear weights, or equivalently $1 \times 1$ convolution, to combine various perturbation maps to create one feature map. When a $1 \times 1$ convolution is applied on the input map, only one element contributes to the one output elements, as opposed to a $3 \times 3$ convolution which involve 9 elements of the input as depicted in Figure 3. Therefore, $1 \times 1$ convolution provides the sparsest interactions possible. Figure 3 shows various commonly practiced convolutions such as (a) the regular convolution, (b) locally connected convolution, (c) tiled convolution, and finally (d) fully connected layer. It is important to note that while a perturbation layer by itself has a receptive field of one pixel, the receptive field of a PNN would typically cover the entire image with an appropriate size and number of pooling layers.

**Parameter sharing**: Although the fixed perturbation masks are shared among different inputs, they are not learnable, and therefore not considered as the parameters in this context. Here, the parameter sharing is carried out again by the $1 \times 1$ convolution that linearly combines the various non-linearly activated perturbation masks.

In addition, PNN has other nice properties such as **multi-scale equivalent convolutions** *i.e.*, adding different amounts of perturbation noise is equivalent to applying convolutions at different scales. More specifically, adding small noise corresponds to applying a small-sized convolutional filter, and adding larger noise corresponds to convolving with a larger filter. Without explicitly setting the filter sizes throughout the network layers, PNN allows the network to adapt to different filter sizes automatically and optimally. Please refer to the supplementary materials for more analysis and discussions. Finally, PNN also has **distance preserving property**. Please also see supplementary for more details.

## 4. Implementation Details

### 4.1. Parameter Savings

The number of learnable parameters in the perturbation layer is significantly lower than those of a standard convolutional layer for the same number of input and output channels. Let the number of input and output channels be $p$ and $q$ respectively. With a convolutional kernel of size of $h \times w$, a standard convolutional layer consists of $p \cdot h \cdot w \cdot q$ learnable parameters. The corresponding perturbation layer consists of $p \cdot m$ fixed perturbation masks and $m \cdot q$ learnable parameters (corresponding to the $1 \times 1$ convolution), where $m$ is the number of fan-out channels of the perturbation layer, and the fan-out ratio $(m/p)$ is essentially the number of perturbation masks applied on each input channel. The $1 \times 1$ convolutions act on the $m$ perturbed maps of the fixed filters to generate the $q$-channel output. The ratio of the number of parameters in CNN and PNN is:

$$\frac{\text{\# param. in CNN}}{\text{\# param. in PNN}} = \frac{p \cdot h \cdot w \cdot q}{m \cdot q} = \frac{p \cdot h \cdot w}{m} \qquad (28)$$

For simplicity, assuming fan-out ratio $m/p = 1$ reduces the parameter ratio to $h \cdot w$. Therefore, numerically, PNN saves $k^2$ parameters during learning for $k \times k$ convolutional filters. Also, PNN allows flexible adjustment of the fan-out ratio to trade-off between efficiency and accuracy.

### 4.2. Learning with Perturbation Layers

Training a network end-to-end with perturbation layers instead of standard convolutional layers is straightforward. The gradients can be back propagated through the $1 \times 1$ convolutional layer and the additive perturbation masks in much the same way as they can be back propagated through standard convolutional layers. Backpropagation through the noise perturbation layer is similar in spirit to propagating gradients through layers without learnable parameters (*e.g.*, ReLU, max pooling, *etc.*). However during learning, only the learnable $1 \times 1$ filters are updated while the additive perturbation masks remain unaffected. For the forward propagation defined in Eq. 4, backpropagation can be computed as:

$$\frac{\partial \hat{\mathbf{y}}}{\partial \mathbf{x}} = \sum_{i=1}^{N} v_i \quad \text{and} \quad \frac{\partial \hat{\mathbf{y}}}{\partial v_i} = \mathbf{x} + \mathbf{n}_i \qquad (29)$$

The perturbation masks are of the same spatial size as the input tensor, and for each input channel, we can generate $m/p$ masks separately ($m/p$ is the fan-out ratio). Specifically, the additive noise in the perturbation masks are independently uniformly distributed. The formulation of PNN does not require the perturbation noise to be a specific type, as long as it is zero-mean and has finite variance. Empirically, we have observed that adding zero-mean Gaussian noise with different variances performs comparably to adding zero-mean uniform noise with different range levels. Since uniform distribution provides better control over the energy level of the noise, our main experiments are carried out by using uniformly distributed noise in the perturbation masks.

## 5. Experiments

### 5.1. ImageNet-1k Classification and Analysis

We evaluate our method on the ImageNet ILSVRC-2012 classification dataset [24] which consists of 1000 classes, with 1.28 million images in the training set and 50k images in the validation set, where we use for testing as commonly practiced. We report the top-1 classification accuracy. All the images are first resized so that the long edge is 256 pixels, and then a $224 \times 224$ crop is randomly sampled from an image or its horizontal flip, with the per-pixel mean subtracted. During testing, we adopt the single center-crop testing protocol. The network architecture we use for this experiment is PNN-ResNet-18 [8], where each standard convolutional layer in a residual unit is replaced by the proposed perturbation layer, as shown in Figure 4.

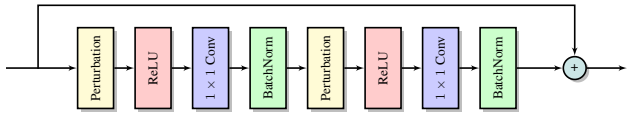

**Figure 4:** Perturbation residual module.

We have experimented with various number of perturbation masks per layer (64, 128, and 256) on the same PNN-ResNet-18 model. The results are consolidated in Table 1 and in Figure 5. As can be seen, compared to the state-of-the-art ResNet-18 performance (single center-crop protocol) on a standard CNN [4, 8], the proposed PNN achieves comparable classification accuracy on ImageNet-1k. It is worth noting that the lightweight design in PNN allows significant parameter savings as well as statistical efficiency compared to the standard CNNs. We have shown the parameter ratio of CNN over PNN in the last column in Table 1. We also present additional experimental results of the PNN-ResNet-50 (with 256 perturbation masks) on ImageNet-1k. The results in Table 2 show that PNN-ResNet-50 performs competitively with the corresponding network with CNN layers [4] on ImageNet-1k.

**Table 1:** Classification accuracy (%) on ImageNet-1k (PNN vs. CNN)

| #Mask | PNN (ResNet-18) | ResNet [4] | Param. Ratio |
|---|---|---|---|
| 256 | 71.84 | 73.27 (34) | 0.9 |
| 128 | 61.74 | 69.57 (18) | 1.8 |
| 64 | 45.92 | 69.57 (18) | 5.9 |

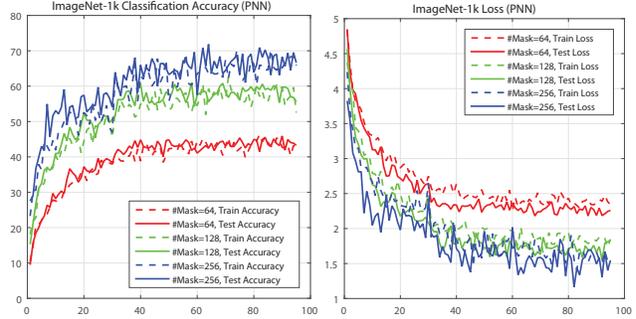

**Figure 5:** Accuracy and loss on ImageNet-1k classification using PNN (ResNet-18) with various number of perturbation masks per layer.

**Table 2:** Classification accuracy (%) on ImageNet-1k (PNN vs. CNN)

| PNN-ResNet-18 | ResNet-18 | PNN-ResNet-50 | ResNet-50 |
|---|---|---|---|
| 71.84 | 69.57 | 76.23 | 75.99 |

### 5.2. ImageNet-100 Classification and Analysis

In this section, we examine the image classification accuracy versus the noise level in the additive perturbation masks. As discussed in the previous section, we are using zero-mean uniform random noise in the perturbation masks, and the noise level $\ell$ here refers to the range of the noise and the PDF is: $f(n_i) = 1/(\ell - (-\ell))$ for $-\ell \leq n_i \leq \ell$.

For faster roll-out, we randomly select 100 classes with the largest number of images (1300 training images in each class, with a total of 130k training images and 5k testing images), and report top-1 accuracy on this ImageNet-100 subset. The network architecture we use for this experiment is PNN-ResNet-50 with 256 perturbation masks per layer. The experimental results are shown in Table 3 as well as in Figure 6. As we observe, adding different amount of perturbation noise does affect classification performance. Moreover, the proposed PNN's performance is similar when the noise level is low, say less than 1. As the noise levels increase, the performance begins to deteriorate. This is expected, as adding too much noise will suppress the useful information carried by the signal itself.

**Table 3:** Classification accuracy (%) on 100-class ImageNet with varying perturbation noise levels. (PNN: ResNet-50, 256 perturbation masks)

| Noise | 0.01 | 0.05 | 0.1 | 0.5 | 1 | 5 |
|---|---|---|---|---|---|---|
| PNN | 81.09 | 81.19 | 81.41 | 81.96 | 76.84 | 60.90 |

### 5.3. CIFAR-10, MNIST Classification and Analysis

In this section, we carry out further classification experiments on CIFAR-10 [13] and MNIST [16] datasets. For both datasets, the initial learning rate is set to $10^{-3}$ and is reduced by a factor of 10 at epoch 60, and then again at epoch 90. The best performing PNN models for each dataset are detailed as follows. For CIFAR-10: PNN-ResNet-50, 64 perturbation masks per layer, and batch size of 10. For MNIST:

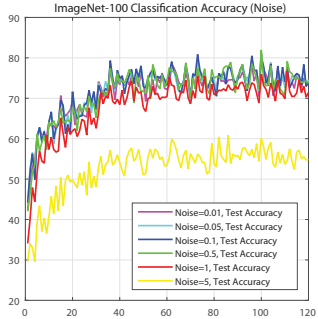

**Figure 6:** Accuracy on ImageNet-100 classification using PNN (ResNet-50, 256 perturbation masks) with various noise levels.

PNN-ResNet-50, 32 perturbation masks per layer, and batch size of 10. Table 4 consolidates the image classification accuracy from our experiments. The best performing PNNs are compared to the state-of-the-art methods as listed in the curated leader board on various image classification tasks [1], as well as several other leading methods such as ResNet [8], Maxout Network [6], Network in Network (NIN) [19], and LBCNN [12]. Our results indicate that PNN is highly competitive with the state-of-the-art results on CIFAR-10.

**Table 4:** Classification accuracy (%) on CIFAR-10 and MNIST. PNN columns only show the best performing model.

|          | PNN   | SoTA [1] | LBCNN | ResNet | Maxout | NIN   |
|----------|-------|----------|-------|--------|--------|-------|
| CIFAR-10 | 94.05 | 96.53    | 92.99 | 93.57  | 90.65  | 91.19 |
| MNIST    | 99.39 | 99.79    | 99.51 | -      | 99.55  | 99.53 |

In addition, Table 5 shows the CIFAR-10 performance on PNN-ResNet-18 model with different number of perturbation masks per layer. As long as the number of perturbation masks is not too few, the network is able to converge and provide competitive classification performance.

**Table 5:** Classification accuracy (%) and ratio of parameters on CIFAR-10 with varying number of perturbation masks. (PNN: ResNet-18, CNN: standard ResNet-18)

| #Mask | 160   | 128   | 96    | 64    | 32    | 16    | 8     |
|-------|-------|-------|-------|-------|-------|-------|-------|
| PNN   | 90.43 | 89.48 | 90.25 | 89.99 | 93.08 | 87.16 | 79.98 |
| Ratio | 1.3   | 2.0   | 3.5   | 7.9   | 31.0  | 120.1 | 451.1 |

The other factor we want to examine is the learning rate. Since we know that learning rate is tightly connected to the batch size from recent findings [7, 28], we vary the batch size as we fix the initial learning rate to be $10^{-3}$. Table 6 shows the CIFAR-10 classification performance with varying batch size as well as varying number of perturbation masks. For this dataset, smaller batch size seems to always work better. This could be due to a batch size of 10 corresponds to the optimal initial learning rate in this case. Also, PNN does not require lots of perturbation masks per layer. Usually, the optimal range is around 48-80 masks per layer for this dataset, and doubling the number of masks from 64 to 128 does not seem to help improve the performance. PNN achieves its best performance on CIFAR-10 with 64 perturbation masks with a ResNet-50 architecture, while the best results for ResNet architecture are obtained with 110 layers, resulting in a $3.1\times$ reduction in parameters while achieving similar performance.

**Table 6:** Classification accuracy (%) on CIFAR-10 with varying batch size and number of perturbation masks. (PNN: ResNet-50)

| #Mask\ Batch Size | 10    | 20    | 40    | 80    |
|-------------------|-------|-------|-------|-------|
| 32                | 90.23 | 87.29 | 83.96 | 79.16 |
| 64                | 94.05 | 90.93 | 88.36 | 85.29 |
| 128               | 93.71 | 90.05 | 88.63 | 85.14 |

### 5.4. Object Detection and Analysis

For this task, we look at the Faster R-CNN model [31] whose region proposal network and detector network both share a common pretrained (on ImageNet) CNN. We study both CNN architectures (VGG-16, ResNet-50) by replacing the convolution layers with the proposed PNN modules. Table 7 shows the mean average precision (mAP) on PASCAL VOC'07 testing set (trained on VOC'07 train+val set, scale=600, batchsize=1, and with ROI align). We observe that PNN-VGG-16 and PNN-ResNet-50 perform comparably to the corresponding network with conv layers [31].

**Table 7:** Detection results (mAP) on PASCAL VOC 2007 testing set.

| PNN-VGG-16 | VGG-16 | PNN-ResNet-50 | ResNet-101 |
|------------|--------|---------------|------------|
| 69.6       | 70.2   | 72.8          | 75.2       |

## 6. Conclusions

Convolutional layers have become the mainstay of state-of-the-art image classification tasks. Many different deep neural network architectures have been proposed building upon convolutional layers, including convolutional layers with small receptive fields, sparse convolutional weights, binary convolutional weights, factorizations of convolutional weights, *etc*. However, the basic premise of a convolutional layer has remained the same through these developments. In this paper, we sought to validate the utility of convolutional layers through a module that is devoid of convolutional weights and only computes weighted linear combinations of non-linear activations of additive noise perturbations of the input. Our experimental evaluations yielded a surprising result, deep neural networks with the perturbation layers perform as well as networks with standard convolutional layers across different scales and difficulty of image classification and detection datasets. Our findings suggest that perhaps high performance deep neural networks for image classification and detection can be designed without convolutional layers.

# References

[1] R. Benenson. Are We There Yet? Classification Datasets Results. http://rodrigob.github.io/are_we_there_yet/build/classification_datasets_results.html. Accessed: 2017-11-15. 8

[2] M. Courbariaux and Y. Bengio. BinaryNet: Training Deep Neural Networks with Weights and Activations Constrained to +1 or -1. *arXiv preprint arXiv:1602.02830*, 2016. 1, 2

[3] M. Courbariaux, Y. Bengio, and J.-P. David. BinaryConnect: Training Deep Neural Networks with Binary Weights During Propagations. In *Advances in Neural Information Processing Systems (NIPS)*, pages 3105–3113, 2015. 1, 2

[4] Facebook. ResNet training in Torch by Facebook. https://github.com/facebook/fb.resnet.torch. Accessed: 2017-11-15. 7

[5] I. Goodfellow, Y. Bengio, and A. Courville. *Deep Learning*. MIT press, 2016. 1, 5

[6] I. J. Goodfellow, D. Warde-Farley, M. Mirza, A. Courville, and Y. Bengio. Maxout Networks. In *30th International Conference on Machine Learning (ICML)*, 2013. 8

[7] P. Goyal, P. Dollár, R. Girshick, P. Noordhuis, L. Wesolowski, A. Kyrola, A. Tulloch, Y. Jia, and K. He. Accurate, Large Minibatch SGD: Training ImageNet in 1 Hour. *arXiv preprint arXiv:1706.02677*, 2017. 8

[8] K. He, X. Zhang, S. Ren, and J. Sun. Deep Residual Learning for Image Recognition. In *Proceedings of the IEEE Conference on Computer Vision and Pattern Recognition (CVPR)*, pages 770–778, 2016. 1, 2, 7, 8

[9] K. He, X. Zhang, S. Ren, and J. Sun. Identity Mappings in Deep Residual Networks. In *European Conference on Computer Vision (ECCV)*, pages 630–645, 2016. 1

[10] A. G. Howard, M. Zhu, B. Chen, D. Kalenichenko, W. Wang, T. Weyand, M. Andreetto, and H. Adam. MobileNets: Efficient Convolutional Neural Networks for Mobile Vision Applications. *arXiv preprint arXiv:1704.04861*, 2017. 1, 2

[11] G. Huang, Z. Liu, K. Q. Weinberger, and L. van der Maaten. Densely Connected Convolutional Networks. *arXiv preprint arXiv:1608.06993*, 2016. 1

[12] F. Juefei-Xu, V. N. Boddeti, and M. Savvides. Local Binary Convolutional Neural Networks. In *Proceedings of the IEEE Conference on Computer Vision and Pattern Recognition (CVPR)*, volume 1, 2017. 2, 3, 8

[13] A. Krizhevsky and G. Hinton. Learning Multiple Layers of Features from Tiny Images. *CIFAR*, 2009. 7

[14] A. Krizhevsky, I. Sutskever, and G. E. Hinton. ImageNet Classification with Deep Convolutional Neural Networks. In *Advances in Neural Information Processing Systems (NIPS)*, pages 1097–1105, 2012. 1, 2

[15] V. Lebedev, Y. Ganin, M. Rakhuba, I. Oseledets, and V. Lempitsky. Speeding-up Convolutional Neural Networks Using Fine-tuned CP-Decomposition. *arXiv preprint arXiv:1412.6553*, 2014. 2

[16] Y. LeCun, L. Bottou, Y. Bengio, and P. Haffner. Gradient-based Learning Applied to Document Recognition. *Proceedings of the IEEE*, 86(11):2278–2324, 1998. 7

[17] Y. LeCun, J. S. Denker, S. A. Solla, R. E. Howard, and L. D. Jackel. Optimal Brain Damage. In *Advances in Neural Information Processing Systems (NIPS)*, volume 2, pages 598–605, 1989. 1

[18] S. Li, J. Park, and P. T. P. Tang. Enabling Sparse Winograd Convolution by Native Pruning. *arXiv preprint arXiv:1702.08597*, 2017. 1, 2

[19] M. Lin, Q. Chen, and S. Yan. Network in Network. In *International Conference on Learning Representations (ICLR)*, 2014. 2, 8

[20] J. Park, S. Li, W. Wen, H. Li, Y. Chen, and P. Dubey. Holistic SparseCNN: Forging the Trident of Accuracy, Speed, and Size. *arXiv preprint arXiv:1608.01409*, 2016. 1, 2

[21] J. Park, S. Li, W. Wen, P. T. P. Tang, H. Li, Y. Chen, and P. Dubey. Faster cnns with direct sparse convolutions and guided pruning. *arXiv preprint arXiv:1608.01409*, 2016. 1, 2

[22] W. H. Press, S. A. Teukolsky, W. T. Vetterling, and B. P. Flannery. Section 2.7.1 Sherman–Morrison Formula. In *Numerical Recipes: The Art of Scientific Computing (3rd ed.)*. Cambridge University Press, New York, 2007. 4

[23] M. Rastegari, V. Ordonez, J. Redmon, and A. Farhadi. XNOR-Net: ImageNet Classification Using Binary Convolutional Neural Networks. In *European Conference on Computer Vision (ECCV)*, pages 525–542, 2016. 1, 2

[24] O. Russakovsky, J. Deng, H. Su, J. Krause, S. Satheesh, S. Ma, Z. Huang, A. Karpathy, A. Khosla, M. Bernstein, et al. ImageNet Large Scale Visual Recognition Challenge. *International Journal of Computer Vision (IJCV)*, 115(3):211–252, 2015. 7

[25] J. Sherman and W. J. Morrison. Adjustment of an Inverse Matrix Corresponding to Changes in the Elements of a Given Column or a Given Row of the Original Matrix (abstract). *Annals of Mathematical Statistics*, 20:621, 1949. 4

[26] J. Sherman and W. J. Morrison. Adjustment of an Inverse Matrix Corresponding to a Change in One Element of a Given Matrix. *Annals of Mathematical Statistics*, 21(1):124–127, 1950. 4

[27] K. Simonyan and A. Zisserman. Very Deep Convolutional Networks for Large-scale Image Recognition. In *International Conference on Learning Representations (ICLR)*, 2015. 1, 2

[28] S. L. Smith, P.-J. Kindermans, and Q. V. Le. Don't Decay the Learning Rate, Increase the Batch Size. *arXiv preprint arXiv:1711.00489*, 2017. 8

[29] J. Sylvester. Sur l'equations en matrices $px = xq$. *C. R. Acad. Sci. Paris*, 99(2):67–71,115–116, 1884. 5

[30] C. Szegedy, W. Liu, Y. Jia, P. Sermanet, S. Reed, D. Anguelov, D. Erhan, V. Vanhoucke, and A. Rabinovich. Going Deeper with Convolutions. In *Proceedings of the IEEE Conference on Computer Vision and Pattern Recognition (CVPR)*, pages 1–9, 2015. 1, 2

[31] J. Yang. A Faster Pytorch Implementation of Faster R-CNN. https://github.com/jwyang/faster-rcnn.pytorch. 8

# Perturbative Neural Networks
## *(Supplementary Material)*

Felix Juefei-Xu
Carnegie Mellon University
felixu@cmu.edu

Vishnu Naresh Boddeti
Michigan State University
vishnu@msu.edu

Marios Savvides
Carnegie Mellon University
msavvid@ri.cmu.edu

This article provides supplementary materials for [1]. In Section 1, we will show the proof of Sherman-Morrison formula. In Section 2, we will show the distance preserving property of PNN. In Section 3, computational and statistical complexity and efficiency of PNN will be discussed. In Section 4, we will discuss the relationship between noise level and convolutional filter size in detail. Finally, in Section 5, some visualization of intermediate responses of both PNN and CNN will be shown.

## 1. Proof of Sherman-Morrison Formula

The Sherman-Morrison formula [6, 7, 4] computes the inverse of the sum of an invertible matrix $A$ and the outer product $uv^\top$ of vector $u$ and $v$. Suppose $A \in \mathbb{R}^{n \times n}$ is an invertible square matrix and $u \in \mathbb{R}^n$, $v \in \mathbb{R}^n$ are column vectors. Then we have $A + uv^\top$ is invertible iff $1 + v^\top A^{-1} u \neq 0$. If $A + uv^\top$ is invertible, then its inverse is given by:

$$(A + uv^\top)^{-1} = A^{-1} - \frac{A^{-1} uv^\top A^{-1}}{1 + v^\top A^{-1} u} \qquad (1)$$

If we already know the inverse of $A$, the Sherman-Morrison formula provides a numerically inexpensive way to compute the inverse of $A$ corrected by $uv^\top$, which in our case, the correction can be seen as a perturbation in PNN. The computation is relatively inexpensive because the inverse of $A + uv^\top$ does not have to be computed from scratch, but can be computed by correcting or perturbing $A^{-1}$.

*Proof:*

($\Rightarrow$) To prove that the forward direction is true, we need to simply verify if the properties of the inverse hold: $XY = YX = I$, where $X = A + uv^\top$, and $Y$ is the RHS of the Sherman-Morrison formula.

First, we verify whether $XY = I$ holds:

$$XY = (A + uv^\top) \left( A^{-1} - \frac{A^{-1} uv^\top A^{-1}}{1 + v^\top A^{-1} u} \right) \qquad (2)$$

$$= AA^{-1} + uv^\top A^{-1} - \frac{AA^{-1} uv^\top A^{-1} + uv^\top A^{-1} uv^\top A^{-1}}{1 + v^\top A^{-1} u} \qquad (3)$$

$$= I + uv^\top A^{-1} - \frac{uv^\top A^{-1} + uv^\top A^{-1} uv^\top A^{-1}}{1 + v^\top A^{-1} u} \qquad (4)$$

$$= I + uv^\top A^{-1} - \frac{u(1 + v^\top A^{-1} u)v^\top A^{-1}}{1 + v^\top A^{-1} u} \qquad (5)$$

$$= I + uv^\top A^{-1} - uv^\top A^{-1} \qquad (6)$$

$$= I \qquad (7)$$

In the same way, we can verify the following:

$$YX = \left( A^{-1} - \frac{A^{-1} uv^\top A^{-1}}{1 + v^\top A^{-1} u} \right) (A + uv^\top) = I \qquad (8)$$

($\Leftarrow$) To prove the backward direction, we suppose that $u \neq 0$, otherwise the result is trivial. Then we can have the following identity:

$$(A + uv^\top)A^{-1}u = u + uv^\top A^{-1}u = (1 + v^\top A^{-1}u)u \tag{9}$$

Since $A + uv^\top$ is assumed to be invertible, it implies that $(A + uv^\top)A^{-1}$ is also invertible (as the product of full rank matrices). So, by our assumption that $u \neq 0$, we have that $(A + uv^\top)A^{-1}u \neq 0$, and by the identity shown above, this means that $(1 + v^\top A^{-1}u)u \neq 0$, and therefore $1 + v^\top A^{-1}u \neq 0$, as was intended to be shown.

## 2. Distance Preserving Property of PNN

PNN also has **distance preserving property**. For input $\mathbf{x}_p$ and $\mathbf{x}_q$, the PNN produces outputs $\hat{\mathbf{y}}_p$ and $\hat{\mathbf{y}}_q$ as follows:

$$\hat{\mathbf{y}}_p = \sum_{i=1}^{N}(\mathbf{x}_q + \mathbf{n}_i)v_i = (\mathbf{x}_p\mathbf{1}^\top + \mathbf{N})\mathbf{v} = \mathbf{x}_p \sum_i^N v_i + \mathbf{N}\mathbf{v}$$

$$\hat{\mathbf{y}}_q = \sum_{i=1}^{N}(\mathbf{x}_q + \mathbf{n}_i)v_i = (\mathbf{x}_p\mathbf{1}^\top + \mathbf{N})\mathbf{v} = \mathbf{x}_q \sum_i^N v_i + \mathbf{N}\mathbf{v}$$

we now examine the distance between $\hat{\mathbf{y}}_p$ and $\hat{\mathbf{y}}_q$ as:

$$\|\hat{\mathbf{y}}_p - \hat{\mathbf{y}}_q\|_2^2 = \left\| \mathbf{x}_p \sum_i^N v_i + \mathbf{N}\mathbf{v} - \mathbf{x}_q \sum_i^N v_i - \mathbf{N}\mathbf{v} \right\|_2^2$$

$$= \left(\sum_i^N v_i\right)^2 \cdot \|\mathbf{x}_p - \mathbf{x}_q\|_2^2 \tag{10}$$

Applying Cauchy-Schwarz inequality, we can have:

$$\|\hat{\mathbf{y}}_p - \hat{\mathbf{y}}_q\|_2^2 \leq N \cdot \|\mathbf{v}\|_2^2 \cdot \|\mathbf{x}_p - \mathbf{x}_q\|_2^2 \tag{11}$$

## 3. Computational and Statistical Complexity

In comparison to standard CNNs, PNNs can be very attractive from the perspective of computational and statistical efficiency.

### 3.1. Computational Efficiency

Computationally, PNNs are significantly more efficient than CNNs, since PNNs completely do away with expensive convolution operations and instead replace them by more efficient weighted linear combinations. Given an input $k \times m \times n$, a convolutional layer with weights $k \times 3 \times 3$ and a corresponding PNN layer with weights $k \times 1 \times 1$, the number of multiplications required by the PNN layer is $= kmn$, while for the CNN needs $\approx 3kmn$ multiplications when using Winograd convolution [2, 8]. In practice, however, we did not observe any speed-up. We believe that this may be since $3 \times 3$ convolutions enjoy highly optimized implementations, while the $1 \times 1$ convolution implementations are not explicitly optimized and typically suffer from cache misses. Storage wise PNN result in smaller models, due to the fewer number of parameters (by a factor of $pq$) in a PNN layer in comparison to a CNN layer with the same number of input and output channels.

### 3.2. Computational Complexity

The weighted linear combination operation of PNN can be implemented either through a $1 \times 1$ convolution with the NCHW data format, or through a fully connected layer with the NHWC data format, with the latter being the more efficient implementation (up to a factor of $2\times$ in our experiments). However in a full network like ResNet, we are constrained to implementing PNN as a $1\times 1$ convolution since the other layers in ResNet are optimized (NVIDIA's cuDNN) only for the NCHW data format. Furthermore, We stress that current cuDNN implementation of conv2d is highly optimized for $3 \times 3$ convolutions with the NCHW data format. We believe that optimized implementations will allow us to realize the full computational benefit of PNN over $3 \times 3$ convolutions. Our current implementation does the following, (1) transforms the

NCHW data into NHWC format, (2) performs a matrix-matrix multiplication (fully connected layer) mapping $p$-channels to $q$-channels, and (3) transforming the data from NHWC back to NCHW. Note that while the fully connected layer is faster than the $1 \times 1$ convolution, transforming the data between the NCHW and NHWC formats is an expensive operation. This new implementation demonstrates a modest 10% speed-up over standard $3 \times 3$ convolution.

### 3.3. Statistical Efficiency

The lower model complexity of PNN makes them an attractive option for learning with low sample complexity. To demonstrate the statistical efficiency of PNN, we perform additional face recognition tasks on the FRGC v2.0 dataset [3] under a limited sample complexity setting. The total number of images in each class ranges from 6 to 132 (51.6 on average). While there are 466 classes in total, we experiment with increasing number of randomly selected classes (10, 50 and 100) with a 60-40, 40-60, and 20-80 train/test split. Across the number of classes and various train/test splits, our network parameters remain the same except for the classification fully connected layer at the end. For both the PNN and CNN we adopt the ResNet-18 architecture with the same number of input and output channels. While the CNN architecture uses $3 \times 3$ filters, our PNN network uses $1 \times 1$ filters, therefore the PNN consists of much fewer learnable parameters ($9\times$ fewer) compared to the CNN. We make a few observations from our experimental results (see Figure 1): (1) PNN converges faster than CNN. This is true across all number of classes as well as all train/test splits. (2) PNN outperforms CNN on this task. Lower model complexity helps PNN prevent over-fitting especially on small to medium-sized datasets.

## 4. Noise Level and Convolutional Filter Size

As we have shown in Section 3.4 of the main paper [1], the difference ($\epsilon_i$) between neighboring pixels can be treated as a random variable with $E(\epsilon_i) = 0$, $E(\epsilon_i^2) = 2\sigma^2(1-\rho) = 2\sigma^2\delta$, and $E(\epsilon_i\epsilon_j) = \sigma^2(1-\rho) = \sigma^2\delta$. Therefore we can further derive the correlation coefficient $\rho_\epsilon$ between $\epsilon_i$ and $\epsilon_j$ as follows:

$$\rho_\epsilon = \frac{E(\epsilon_i\epsilon_j)}{\sqrt{E(\epsilon_i^2)}\sqrt{E(\epsilon_j^2)}} = \frac{\sigma^2\delta}{\sqrt{2\sigma^2\delta}\sqrt{2\sigma^2\delta}} = \frac{\sigma^2\delta}{2\sigma^2\delta} = \frac{1}{2} \qquad (12)$$

The derivation in Section 3.4 suggests that the choice of distribution of the noise $\epsilon_i$ in the PNN formulation can potentially be flexible, as long as its first and second order statistics behave according to the expressions shown in the main paper [1].

Next, we will discuss how the additive perturbation noise level (variance of the noise) in PNN can be related to the size of the convolutional filter in traditional CNN. More specifically, adding noise with low variance is equivalent to using a small-sized convolutional filter, while adding noise with larger variance is equivalent to convolving with a larger filter. Therefore, by adjusting the noise distribution (variance) PNN can potentially mimic a CNN with different filter sizes using different noise variance in each channel of the perturbation noise layer without having to explicitly decide on the filter sizes throughout the network layers.

Let us revisit the equivalent noise, $\sum_{i\in\mathbb{N}_c} \epsilon_i w_i$, in a CNN formulation in Eq. 24 of the main paper [1], or equivalently $\sum_{i\in\mathbb{N}_c} \epsilon_i w_i'$ with a normalization term divided on both sides of the equation. It can be seen that in Eq. 24 as a larger-sized convolutional filter is applied, more $\epsilon_i w_i$ terms are involved in the equivalent noise expression.

Let us assume that random variables $\epsilon_i$'s are Gaussian and according to the aforementioned discussion, they are not independent, with correlation coefficient $\rho_\epsilon$. The sum of two non i.i.d. Gaussian variables $\epsilon_i + \epsilon_j \sim \mathcal{N}(0, E(\epsilon_i^2) + E(\epsilon_j^2) + 2E(\epsilon_i\epsilon_j)) \sim \mathcal{N}(0, 2\sigma^2\delta + 2\sigma^2\delta + 2\sigma^2\delta)$. An extension to this is that the linear combination of two non i.i.d. Gaussian variables $w_i\epsilon_i + w_j\epsilon_j \sim \mathcal{N}(0, w_i^2 E(\epsilon_i^2) + w_j^2 E(\epsilon_j^2) + 2w_i w_j E(\epsilon_i\epsilon_j)) \sim \mathcal{N}(0, w_i^2 2\sigma^2\delta + w_j^2 2\sigma^2\delta + 2w_j w_j \sigma^2\delta)$. Therefore, adding up more terms as the convolutional filter size becomes larger would result in a Gaussian distributed perturbation noise with larger variance.

The above analysis is also verified empirically using random convolutional filters of various sizes. We have randomly collected 500 images from randomly chosen 100 classes from ImageNet [5]. We have also generated 100 uniformly distributed random convolutional filters of sizes $3 \times 3$, $5 \times 5$, and $7 \times 7$ each. Images and convolutional filters are single-channeled for simplicity. We have convolved the 500 images with each set of 100 convolutional filters and study the statistics of the equivalent perturbation noise maps. The statistics are shown in Table 1 and the histograms are shown in Figure 2. As can be seen, empirically, larger filter size does correspond to larger noise level being added.

## 5. Visualizing Intermediate Responses

Finally, we provide a visual comparison (see Figure 3) of the intermediate feature maps of CNN and PNN for a ResNet-18 architecture trained on the ImageNet dataset. We observe that in the initial layers, the feature maps of the CNN and PNN both capture similar global image statistics. However, at higher layers (last column of Figure 3) the feature maps of the PNN are sharper than those of the CNN. Despite this difference, both the networks (PNN and CNN) exhibit similar performance on the ImageNet dataset. Even though the noise perturbation layer proposed in this paper does not explicitly aggregate local statistics, as a convolutional layer does, the PNN network still operates on local statistics computed through the pooling layers. Therefore, with enough number of pooling layers the higher layers of the PNN network effectively have a receptive field that spans the entire input image. We do note that the receptive field of PNN is still smaller than the equivalent CNN architecture.

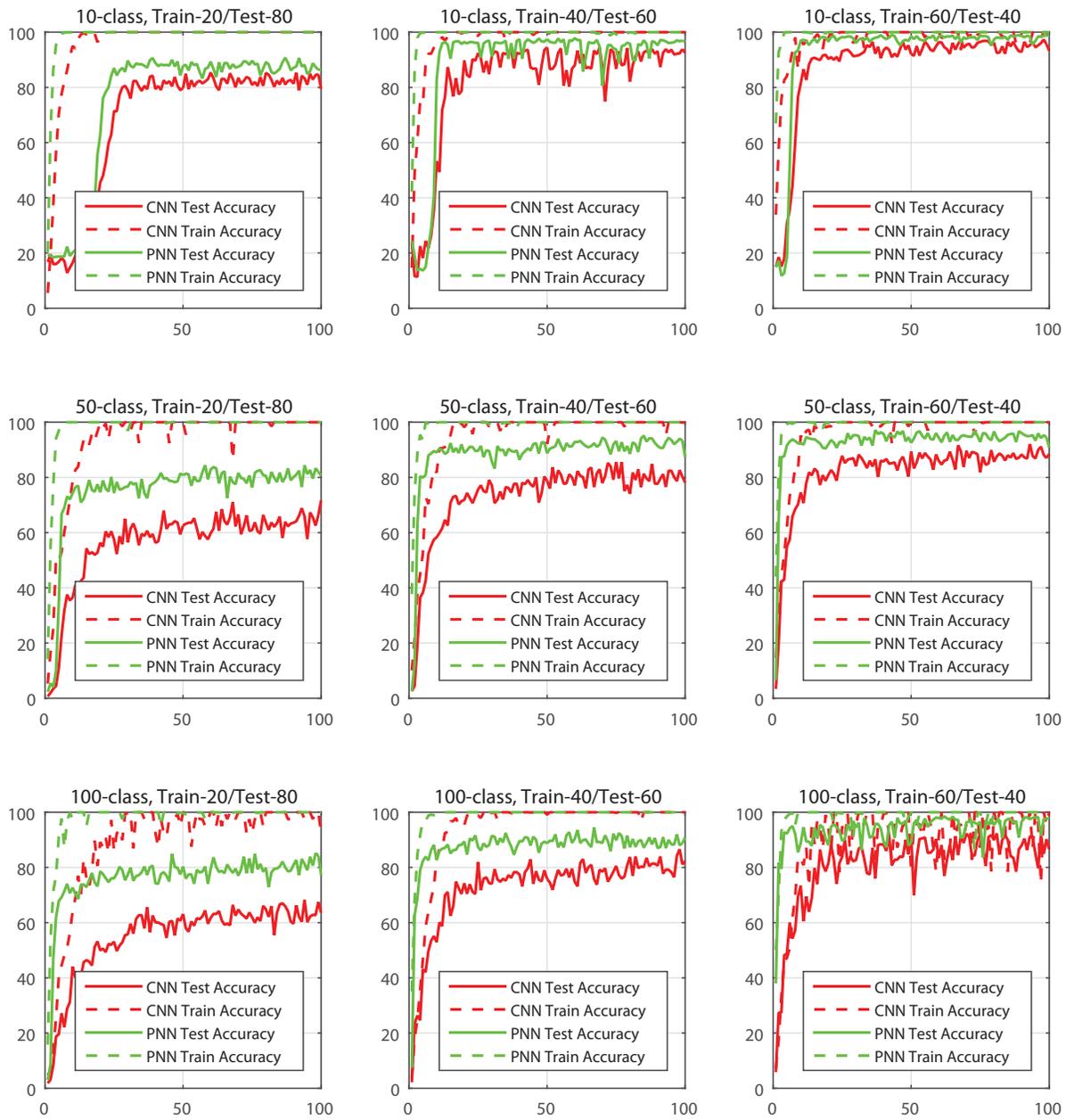

**Figure 1:** Classification on FRGC dataset with low sample complexity. Rows 1-3: 10-class, 50-class, and 100-class classification problems. Columns 1-3: 20-80, 40-60, and 60-40 train/test splits. In general, PNN shows statistical efficiency over CNN.

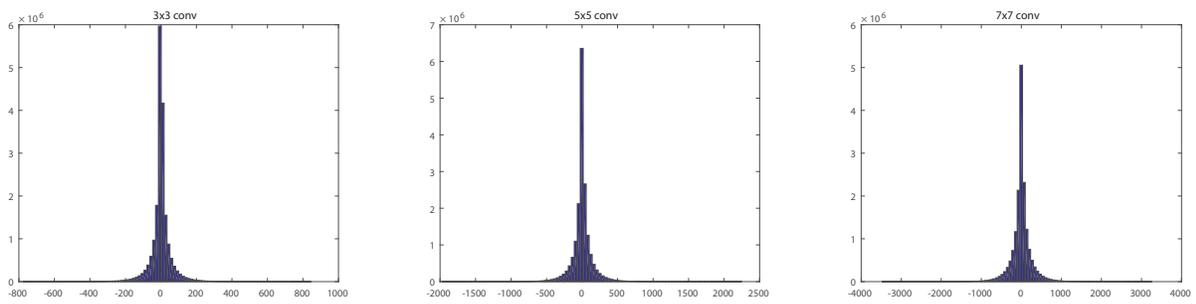

**Figure 2:** Histograms of the difference maps for various convolutional filter sizes. Please note the different x-axis limits.

Table 1: Statistics on the difference maps for various convolutional filter sizes.

|      | $3 \times 3$ | $5 \times 5$ | $7 \times 7$ |
|------|--------------|--------------|--------------|
| Mean | 0.1840       | 0.4115       | -0.2746      |
| Std. | 57.0748      | 141.2023     | 246.2841     |

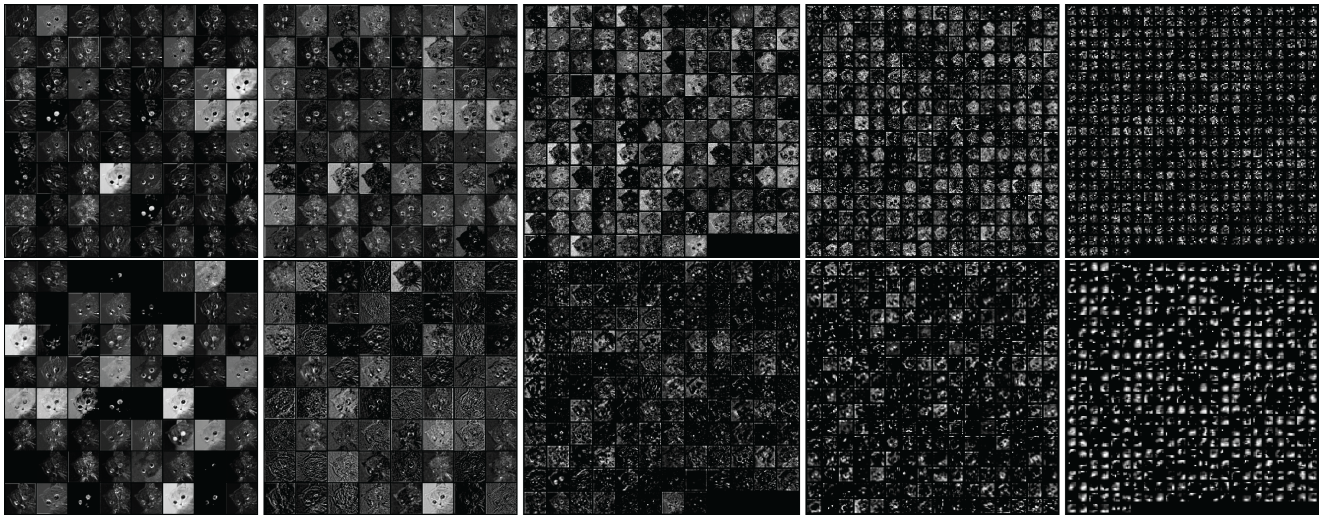

**Figure 3:** Feature maps of the same image from different layers of 18-layer ResNet architecture trained on the ImageNet dataset. The top row corresponds to PNN, while the bottom row corresponds to CNN.

## References

[1] F. Juefei-Xu, V. N. Boddeti, and M. Savvides. Perturbative Neural Networks. In *Proceedings of the IEEE Conference on Computer Vision and Pattern Recognition (CVPR)*, volume 1, 2018. 1, 3

[2] A. Lavin and S. Gray. Fast Algorithms for Convolutional Neural Networks. In *Proceedings of the IEEE Conference on Computer Vision and Pattern Recognition*, pages 4013–4021, 2016. 2

[3] P. J. Phillips, P. J. Flynn, T. Scruggs, K. W. Bowyer, J. Chang, K. Hoffman, J. Marques, J. Min, and W. Worek. Overview of the Face Recognition Grand Challenge. In *Proceedings of the IEEE Conference on Computer Vision and Pattern Recognition (CVPR)*, volume 1, pages 947–954, 2005. 3

[4] W. H. Press, S. A. Teukolsky, W. T. Vetterling, and B. P. Flannery. Section 2.7.1 Sherman–Morrison Formula. In *Numerical Recipes: The Art of Scientific Computing (3rd ed.)*. Cambridge University Press, New York, 2007. 1

[5] O. Russakovsky, J. Deng, H. Su, J. Krause, S. Satheesh, S. Ma, Z. Huang, A. Karpathy, A. Khosla, M. Bernstein, et al. ImageNet Large Scale Visual Recognition Challenge. *International Journal of Computer Vision (IJCV)*, 115(3):211–252, 2015. 3

[6] J. Sherman and W. J. Morrison. Adjustment of an Inverse Matrix Corresponding to Changes in the Elements of a Given Column or a Given Row of the Original Matrix (abstract). *Annals of Mathematical Statistics*, 20:621, 1949. 1

[7] J. Sherman and W. J. Morrison. Adjustment of an Inverse Matrix Corresponding to a Change in One Element of a Given Matrix. *Annals of Mathematical Statistics*, 21(1):124–127, 1950. 1

[8] S. Winograd. On Multiplication of Polynomials Modulo a Polynomial. *SIAM Journal on Computing*, 9(2):225–229, 1980. 2